\pdfoutput=1

\documentclass{article}
\usepackage{spconf,amsmath,graphicx}

\usepackage{hyperref}
\usepackage{cleveref}

\usepackage{makecell}
\usepackage{graphicx}
\usepackage{multirow}
\usepackage{float}
\usepackage{caption}
\usepackage{array}
\usepackage{url}
\usepackage[numbers,sort&compress]{natbib}

\title{Eye Motion Matters for 3D Face Reconstruction}
\name{Xuan Wang, Mengyuan Liu $^\dag$\thanks {$^\dag$Corresponding author.\\
\indent \indent This work was supported by National Natural Science Foundation of China (No. 62203476), Natural Science Foundation of Shenzhen (No. JCYJ20230807120801002).
}
}

\address{National Key Laboratory of General Artificial Intelligence, Peking University, Shenzhen Graduate School \\
\indent
\url{wxuan2060@gmail.com}, \ \url{nkliuyifang@gmail.com}
}

\begin{document}
%
\maketitle
%
\begin{abstract}
Recent advances in single-image 3D face reconstruction have shown remarkable progress in various applications. Nevertheless, prevailing techniques tend to prioritize the global facial contour and expression, often neglecting the nuanced dynamics of the eye region. In response, we introduce an Eye Landmark Adjustment Module, complemented by a Local Dynamic Loss, designed to capture the dynamic features of the eyes area. Our module allows for flexible adjustment of landmarks, resulting in accurate recreation of various eye states. In this paper, we present a comprehensive evaluation of our approach, conducting extensive experiments on two datasets. The results underscore the superior performance of our approach, highlighting its significant contributions in addressing this particular challenge.
\end{abstract}
\section{Introduction}
\label{sec:intro}
\vspace{-0.5em}
Digital humans has garnered substantial attention in recent years. The development of digital humans necessitates the integration of several foundational technologies, like human body analysis  \cite{tu2019action, liu2022generalized, tu2022joint, zhang2022zoom, liu2023temporal, liu2023novel}, 3D face analysis \cite{spl_3dface_alignment1, spl_3dface_alignment2} and so on. As a core component, 3D face reconstruction has a significant positive impact on various applications, including face animation, face reenactment and virtual makeup \cite{22survey}. 

3D face reconstruction task aims to transform 2D image into a 3D facial model. This process goes beyond merely replicating the face's physical appearance and textural details; it also accurately reflects subtle expressions and dynamic changes, thus offering a lifelike and comprehensive digital representation.

During the last two decades, monocular 3D face reconstruction has made significant progress in the context of deep learning. This evolution has transitioned from time-consuming and ill-optimized analysis-by-synthesis methods \cite{3dmm, optima1, optima2}, to current CNN-based techniques \cite{spl_3dface_fitting1, spl_3dface_fitting2, mofa, 3ddfa_v2, 3dmm_cnn, mica, deep3d}. MoFA \cite{mofa} proposed a deep convolutional auto-encoder trained on very large unlabeled data. In Deep3DFaceRecon \cite{deep3d}, a robust and hybrid loss function was leveraged in a weakly supervised manner, to combine low-level and perception-level face information. Meanwhile, the recent MICA \cite{mica} focused on a supervised training scheme, taking advantage of face recognition network to yield a robust 3D face shape estimator. Furthermore, the pursuit of facial details, like expressions and lips, has garnered significant attention. For example, DECA \cite{deca} employed a detail-consistency loss to separate individual specific details from expression dependencies. EMOCA \cite{emoca} proposed an emotion-aware consistency loss to resolve mismatched emotional content between facial geometry and the input image. Additionally, SPECTRE \cite{lipread} revolutionized 3D face reconstruction from video data, introducing a ``lipread" loss to prioritize speech-related facial dynamics.

\begin{figure}
    \centering
    \centerline{\includegraphics[width=\columnwidth]{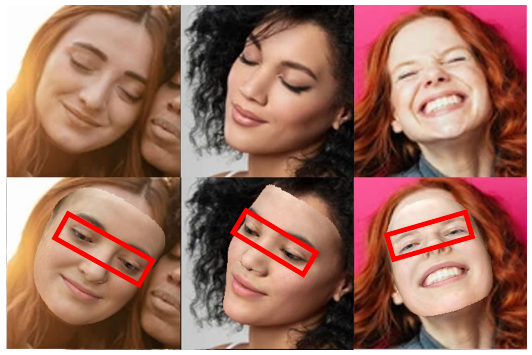}}
    \vspace{-0.8em}
    \caption{\textbf{Current reconstruction problem of closed-eyes images.} The photos are coming from Deep3DFaceRecon \cite{deep3d}.}
    \label{fig:motivation}
    \vspace{-1em}
\end{figure}

The aforementioned CNN-based models have undeniably made substantial strides in capturing facial identity, facilitating robust 3D head reconstructions. However, when these methods are applied to face images characterized by distinct eye dynamics, a notable limitation emerges— the generated reconstructions invariably depict eyes in a consistently open state, as seen in Fig. \ref{fig:motivation}. This observation underscores a critical gap in current research efforts, that the rich dynamic information encapsulated within the eye region remains underutilized. 

To address the problem of eye state inconsistency, we propose an pioneering solution encompassing the Eye Landmark Adjustment Module (ELAM) and a Local Dynamic Loss (LDL) to guide the fitting. ELAM serves as the key to mitigate landmark inaccuracies due to the relatively small eye area of the entire face. Additionally, we propose LDL that leverages the weighted relative distance between the upper and lower eyelids or the lips. This novel loss mechanism further enhances the model's ability to adapt to varying dynamics of small regions. The contributions of our work can be summarized as follows:
\begin{itemize}
    \vspace{-1em}
    \item  We first propose, to the best of our knowledge, an Eye Landmark Adjustment Module, which can guarantee correct eye landmark information.
    \vspace{-1em}
    \item We propose a Local Dynamic Loss to accurately reconstruct relatively small areas of the entire face, bringing positive improvement to 3D face reconstruction.
    \vspace{-1em}
    \item We introduce the use of accuracy and F1-score as criteria for eye region reconstruction.
    \vspace{-1em}
    \item The code will be open-sourced at \url{https://github.com/WangXuan2401/3DFaceRecon}.
\end{itemize}

\begin{figure}
    \centering
    \centerline{\includegraphics[width=\columnwidth]{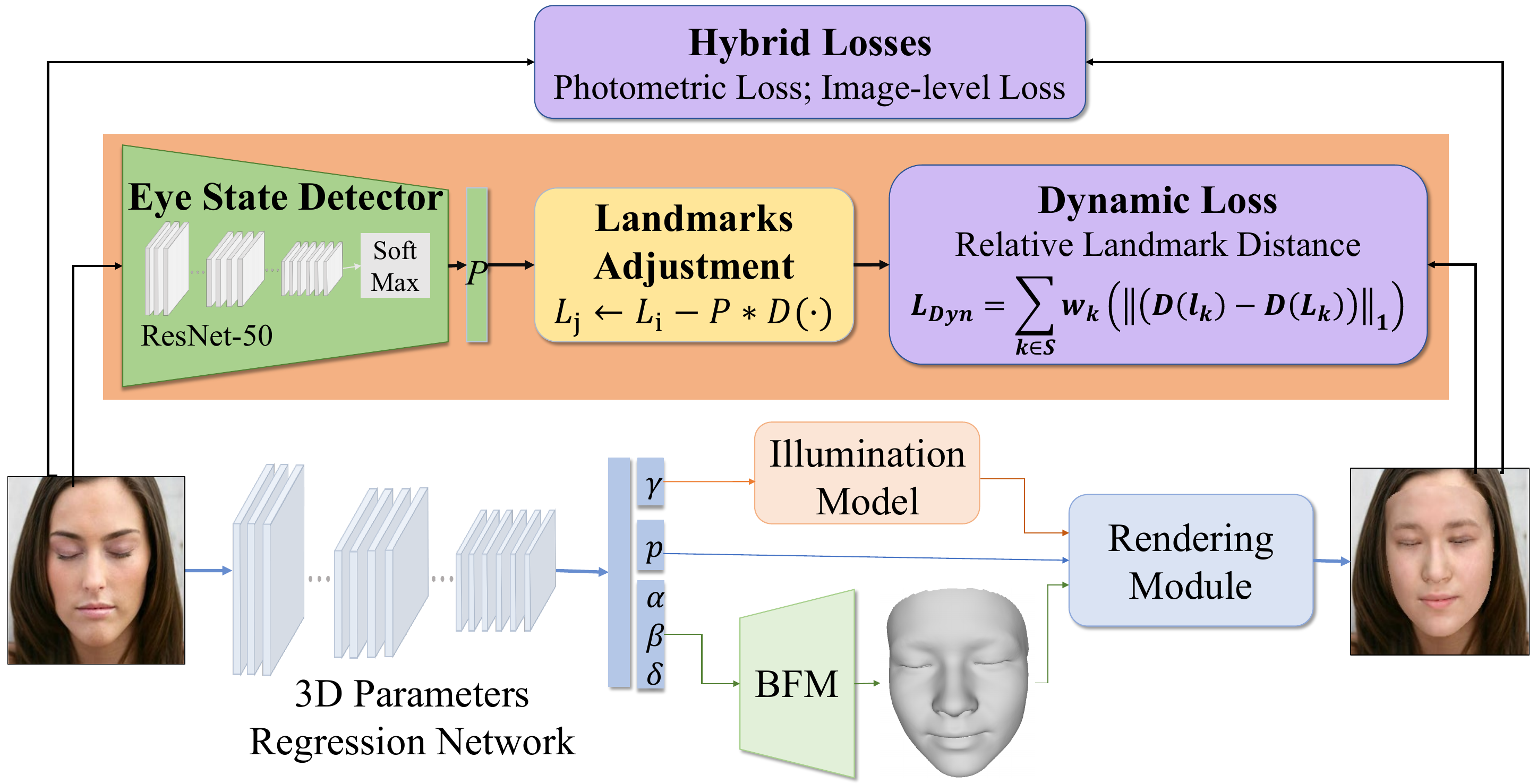}}
    \vspace{-0.8em}
    \caption{\textbf{Overview of our 3D face reconstruction model.} As highlighted in the centre, we introduced ELAM and LDL to effectively address the eye motion issue for 3D face reconstruction. The lower section of the process achieved the basic 3D mesh, while the top layer, consisting of a hybrid loss function, ensures robust face contours and skin textures.}
    \label{fig:framework}
    \vspace{-1em}
\end{figure}

\section{Methodology}
\label{sec:method}

\subsection{Overall Famework}
Fig. \ref{fig:framework} illustrates the overall framework of our 3D face reconstruction model, which comprises three core components: basic models, ELAM and LDL. In order to accurately capture the intricate features of facial images, it is necessary to incorporate a wide array of parameters originating from the 3D statistical model, camera model, and lighting model. Furthermore, to fill the gap in current research on dynamic analysis of the eye region, we develop a new method called ELAM. By integrating an eye state detector and flexible landmarks adjustment rule, ELAM is able to catch subtle changes in eyes. During the training stage, we propose LDL to capture the information of the eye and mouth regions, combined with a hybrid loss function to capture the overall face features.

\vspace{-0.4em}
\subsection{Basic Models}
3D Morphable Model (3DMM) \cite{3dmm} was designed to capture prior knowledge for object presentation. After combing shape and texture parameters from the BFM model \cite{bfm2009} with expression bases from the FaceWarehouse dataset \cite{facewarehouse}, the output 3D face mesh can be represented as follows:
\begin{equation}
    S(\alpha, \beta) = \textbf{S} + B_{S}\alpha + B_{E}\beta
\end{equation}
\begin{equation}
    T(\delta) = \textbf{T} + B_T\delta
\end{equation}
where $\textbf{S}$, $\textbf{T}$ are the average shape and texture of the database faces. $B_S$, $B_E$ and $B_T$ denote the PCA vector bases. The coefficient $\alpha \in R^{\left| \alpha \right|}$, $\beta \in R^{\left| \beta \right|}$ and $\delta \in R^{\left| \delta \right|}$.

A camera model describes the imaging geometry, in particular how positions $ \textbf{v} = [u,v,w]^T$ in the 3D world are projected onto 2D image plane $\textbf{x} = [x,y]^T$. We employ the commonly used perspective camera model for projection. And pose parameters can be represented by camera extrinsic in terms of rotation $R \in SO(3)$ and translation $T\in R^3$.

During the image formation stage, illumination and reflectivity must be modelled as they greatly affect facial appearance, especially skin. The spherical harmonic lighting model \cite{sh} can provide a better approximation of complex natural lighting. Therefore, it is used to accurately express lighting information $\gamma \in R^{\left|\gamma\right|}$.

Through the above introduction to the three basic models, we can explain the 3D face through equation (3):
\vspace{-0.35em}
\begin{equation}
    X = (\alpha, \beta, \delta, \gamma, p) \in R^{\left|X\right|}
    \vspace{-0.35em}
\end{equation}

$\alpha, \beta, \delta$ are the coefficients corresponding to shape, texture and expression. $\gamma$ represents the illumination information. $p$ is the face pose. Based on previous experience \cite{deep3d, deca, emoca, lipread}, we also use ResNet-50 \cite{resnet50} to regress these parameters.

\subsection{ELAM: Eye Landmarks Adjustment Module}
3D face reconstruction based on 3DMM is always sensitive to initial conditions, especially the detected landmarks. Given the relatively diminutive portion of the face occupied by the eyes, the accurate determination of their positions become more challenging. So we propose ELAM, including an eye state detector and a flexible landmarks adjustment rule.

Integrating an eye state detector can increase the reliability of landmark detection, which is often prone to errors. In our work, we use ResNet-50 as such a detector, pretrained on the CEW \cite{cew} sub-dataset, achieving testing accuracy of 97.31\% with a loss of 0.0015. The number of output neurons in the fully connected layer is two. During the prediction stage, we use the SoftMax function within the output layer to calculate the probability, and set the threshold as 0.5. The integration of the eye state detector significantly enhances the precision of landmark detection within the eye region.

To account for variations in eye state, we develop a scheme to flexibly adjust the position of erroneous landmarks. The probability $P$ obtained from the eye state detector representing varying degree of eye closure, which serves as a pivotal component, endowing our approach with flexibility. For instance, when $P=0.5$, it signifies a likelihood of the eye being in a semi-closed state, implying that the upper eyelid should ideally be positioned midway along the y-axis direction of the entire eye. The existence of $P$ greatly broadens the range of eye-level input images, making our method more adaptable to diverse dynamic scenarios. Therefore, we can summarize the flexible eye landmark adjustment rule as follows: for erroneous landmark positions within the eye region, For erroneous eye region landmarks, we adjust the upper eyelid's y-axis position based on `P', adapting precisely to different eye states. The formulation of this rule, shown in (4), provides a better understanding.
\begin{equation}
    L_i \leftarrow L_i-P |L_i-L_j |
\end{equation}
where $L_i$ and $L_j$ denote the y-axis position of upper and lower eyelid landmarks, respectively. $i \in \{37, 38, 43, 44\}$, $j \in \{41, 40, 47, 46\}$ are the exact indexes of our model, and there is a one-to-one correspondence between the two sets.

\begin{table}[t]
\small
\setlength{\tabcolsep}{6pt}
\renewcommand{\arraystretch}{1.35}
\caption {\textbf{Ablation study on ELAM.} Accuracy and F1 are used to measure the reconstruction quality on CEW dataset, where `w/o' means without and `w' represents with.}
\vspace{-0.6em}
\begin{tabular}{m{74pt}<{\centering}|m{68pt}<{\centering}|m{68pt}<{\centering}}
\hline
    Type& Acc&  F1 \\
\hline
    w/o ELAM & 51.7\% & 68.1\% \\
    \cline{1-3}
    w/ ELAM & \textbf{96.6\% (+44\%)} & \textbf{96.7\% (+28\%)} \\
\hline
\end{tabular}
\vspace{-1em}
\label{table1}
\end{table}

\vspace{-0.5em}
\subsection{LDL: Local Dynamic Loss}
Given the challenge of reconstructing small facial areas like the eyes and mouth, we introduce the Local Dynamic Loss. This loss function leverages the relative distances between upper and lower landmark pairs within these specific regions, facilitating the meticulous capture of critical information vital to their accurate reconstruction.

LDL operates on landmark point pairs representing the upper and lower eyelids $(e_i,e_i')$ and lip pairs $(m_j,m_j')$. By measuring the disparity between $D(e_i,e_{i'})$ and $D(E_i,E_{i'})$ - distance value of landmark pairs in 2D image plane - a more precise reproduction of the eyes and mouth in varying states can be attained. Notably, the distance value here is a relative to a specific local area, distinct from the absolute landmark distance value in the following hybrid loss function. For each landmark pair k, We apply an optional weight $w_k$ from the set {0, 1} to specific areas, functioning similarly to a mask. The below formula outlines the function of LDL:
\begin{equation}
    LDL = \sum _{k \in S} w_k \| { D_{l_k} - D_{L_k} }\| _1
    \vspace{-0.2em}
\end{equation}
where S is the set of keypoint pairs. D(·) is the relative distance calculation function. $ l_k$ represents the key pair of the reconstructed 3D landmark vertices projected onto the image plane, which can be considered as the ``predicted value", while $L_k$ is interpreted as``ground truth".

In addition to the above LDL, we also employ a hybrid loss inspired by Deng et al. \cite{deep3d}, better combining image-level and perceptual-level features. Image-level information includes a photo loss computed using facial skin masks and a landmark loss computed based on absolute distances. The perceptual loss quantifies the cosine distance of deep features. The following formula (6-7) shows the content of the hybrid loss function. For more details, see \cite{deep3d}.
\begin{equation} 
    \vspace{-0.4em}
    L_{img} = \frac{\sum_{i \in M} A_i \cdot  \| {I_i -I'_i}\|_2} {\sum_{i \in M} A_i} + \sum_{n=1}^{N} w_n  \| {q_n -q'_n}\|_1
    \vspace{-0.3em}
\end{equation}

\begin{equation}
    L_{per} = 1 - \frac{<f(I), f(I')>}{\|f(I)\| \cdot \|f(I')\|}
    \vspace{-0.2em}
\end{equation}
Among them, $i$ denotes the pixel index, $M$ is the reprojected face region, and $A$ is attention mask. $w_n$ is the given weight, 1 is set for most keypotins while 1.5 is set for eye and mouth regions in our experiment. f($\cdot$) represents the deep feature encoding from Arcface \cite{arcface}, while $<\cdot, \cdot>$ is the inner product.

In summary, the final loss function is a weighted sum of the local dynamic loss and the hybrid loss function. Specifically, the weights are set as $w_{dyn}$ = 0.5, $w_{img}$ = 1.8, and $w_{per}$ = 0.17 respectively.

\begin{table}[t]
\small
\setlength{\tabcolsep}{4pt}
\renewcommand{\arraystretch}{1.35}
\caption{\textbf{Ablation study on LDL.} The NoW Benchmark error quantifies the impact of varied weighted local dynamic losses. Here, `(+mouth)' considers only the mouth area, `(+eye, +mouth)' involves both eye and mouth regions, and `w/o' and `w' denote `without' and `with', respectively.}
\vspace{-0.6em}
\begin{tabular}{m{78pt}<{\centering}|m{50pt}<{\centering}|m{45pt}<{\centering}|m{40pt}<{\centering}}
\hline
    Type& Median(mm)&  Mean(mm) & Std\\
\hline
    w/o $LDL$ & 1.698 & 2.278 & 2.415 \\
    \cline{1-4}
    w/ $LDL$ (+mouth) & 1.505 & 2.050 & 2.322 \\
    \cline{1-4} 
    \rule{0pt}{16pt}
    \makecell{w/ $LDL$ \\ (+eye, +mouth)} & 
    \rule{0pt}{16pt}
    \textbf{\makecell{1.487 \\ (-0.21)}} & 
    \rule{0pt}{16pt}
    \textbf{\makecell{2.033 \\ (-0.24)}} & 
    \rule{0pt}{16pt}
    \textbf{\makecell{2.297 \\ (-0.12)}} \\
    
\hline
\end{tabular}
\label{table2}
\vspace{-0.5em}
\end{table}

\begin{figure}[t]
    \centering
    \centerline{\includegraphics[width=\columnwidth]{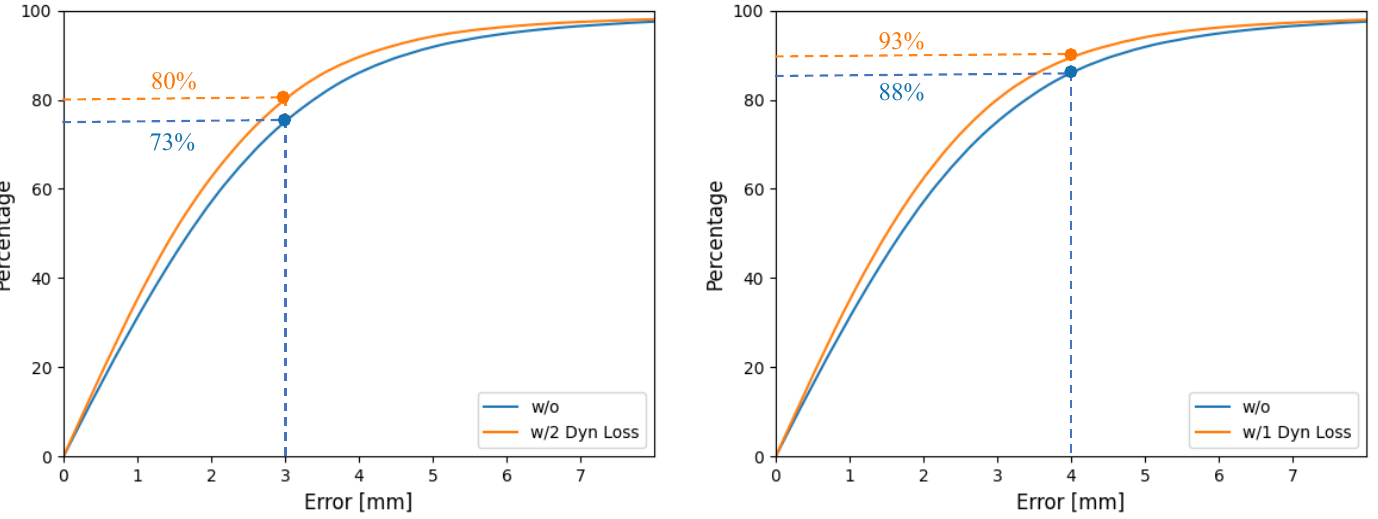}}
    \vspace{-0.5em}
    \caption{\textbf{Comparison Results on Cumulative Errors.}}
    \label{dyn_loss}
    \vspace{-1.5em}
\end{figure}

\begin{table}[t]
\caption{\textbf{Performance Comparison on NoW Benchmark.}}
\vspace{-0.6em}
\label{table3}
\small
\setlength{\tabcolsep}{6pt}
\setlength{\tabcolsep}{6pt}
\renewcommand{\arraystretch}{1.35}
\begin{tabular}{m{50pt}<{\centering}|m{50pt}<{\centering}|m{50pt}<{\centering}|m{50pt}<{\centering}}

\hline
    Method& Median(mm)&  Mean(mm) & Std\\
\hline
    PRNet & 1.499 & 1.976 & 1.881 \\
    \cline{1-4}
    UMDFA & 1.515 & 1.891 & \textbf{1.570} \\ 
    \cline{1-4}
    3DMM-CNN & 1.845 & 2.330 & 2.046 \\
    \cline{1-4}
    \textbf{ours} & \textbf{1.336} & \textbf{1.848} & 2.291 \\
\hline
\end{tabular}
\vspace{-0.3em}
\end{table}  

\begin{figure}[t]
    \centering
    \centerline{\includegraphics[width=0.45\columnwidth]{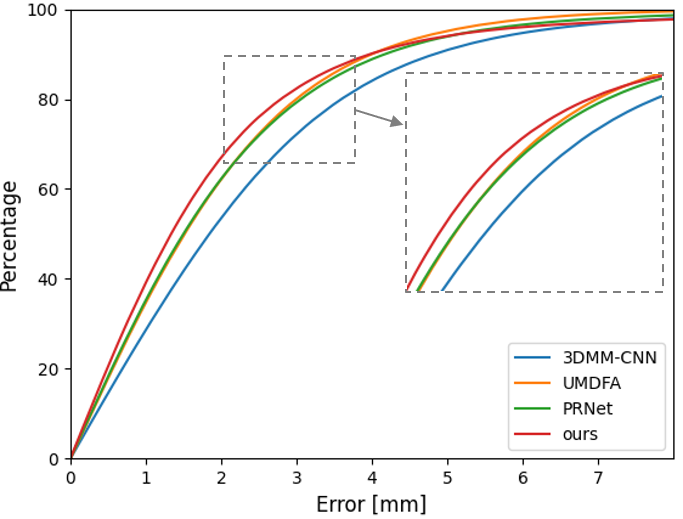}}
    \vspace{-0.4em}
    \caption{\textbf{Cumulative Errors from NoW Benchmark.}}
    \label{comparison}
\end{figure}

\section{Experiment}
\label{sec:experiment}
\vspace{-0.2em}
\subsection{Dataset and Settings}
FFHQ [33] is a dataset with high-quality face images taken in the wild with different lighting and poses. The dataset is also diverse in age, gender and occlusion. In our work, we select about 15k images for training. Furthermore, we also use the CEW dataset \cite{cew} in our experiments, to better highlight the reconstruction effect of our model on eye dynamic features. This dataset includes a total of 2k face images with two different states of eyes open and closed.

All models in this paper are implemented on the PyTorch framework. The core structure is a ResNet-50 pre-trained network for 3D face parameters regression, which is obtained by training on ImageNet. The initial learning rate of the network is 1e-4. And 20 epochs are performed using an Adam optimizer with a batch size of 5. The number of regression parameters in ResNet-50 is 239.

\vspace{-0.5em}
\subsection{Metrics} 
We use shape errors from the NoW Benchmark \cite{now_challenge} to better evaluate the differences between the generated and the scanned 3D facial meshes. Specifically, the errors are obtained by calculating the distance between each vertex of the generated face mesh and the nearest vertex of the ground-truth. In our work, the number of all vertices on a face is 35709. Further details can be found in \cite{now_challenge}.


Additionally, to better evaluate the ability of our method to accurately reconstruct 3D faces under different eye states, we propose to use the accuracy and F1 scores evaluated on face images. In this study, a reconstruction is deemed `true' if it matches the input image's eye state, i.e., a completely closed eye in the input leading to a fully closed eye in the reconstruction, and similarly for open eyes. Conversely, any deviation is classified as `false'.

\vspace{-0.5em}
\subsection{Ablation Studies}
From the Table \ref{table1}, we can find our Eye Landmark Adjustment Module brings a great improvement, reaching accuracy and F1 score by 97\%. The lack of the module would impede the keypoints detection of dynamic eyes region, leading to a inaccurate 3D face reconstruction. Here, the open state is defined as positive at the time of assessment.

Table. \ref{table2} and Fig. \ref{dyn_loss} present the performance with and without the Local Dynamic Loss function tested on the NoW Benchmark. 
When the dynamic loss function is applied to the eye and mouth areas, the error between the reconstructed face and the real human face scan is reduced by approximately 0.2 mm. This indicates that changes in the local dynamic regions still have a noticeable negative impact on the reconstructed 3D face. Concurrently, this also further validates the effectiveness of our local dynamic loss function.

\vspace{-0.5em}

\subsection{Overall Comparison}
\begin{figure}[t]
    \centering
    \centerline{\includegraphics[width=0.95\columnwidth]{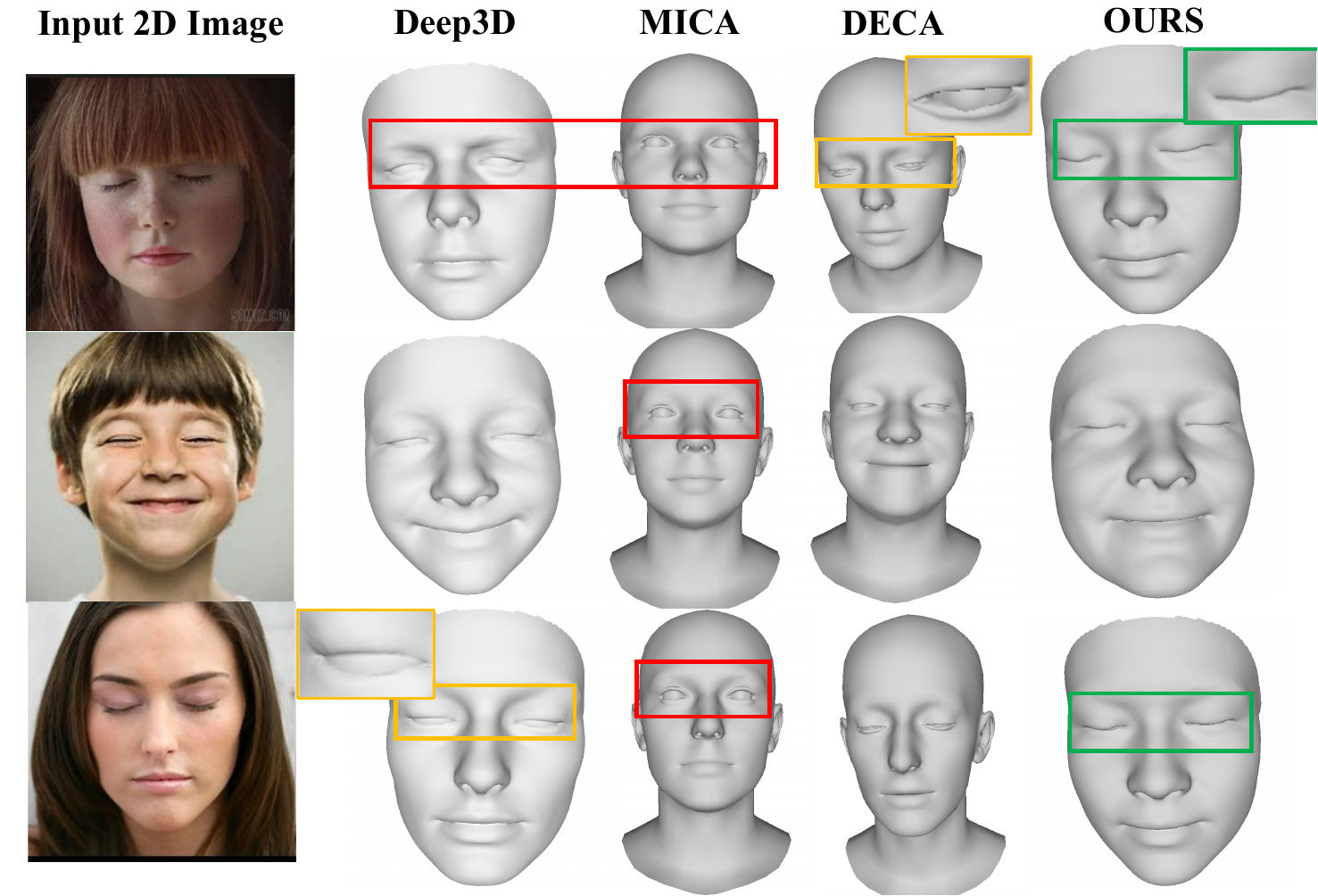}}
    \caption{\textbf{Comparison Results on some closed-eye images.}}
    \label{fig:comparison}\vspace{-1em}
\end{figure}

Fig. \ref{fig:comparison} compares our proposed model with others, including Deep3d \cite{deep3d}, DECA \cite{deca} and MICA \cite{mica}. The red boxes highlight erroneous in the eye region, while the yellow boxes depict slightly improved reconstruction, particularly for DECA. The green box represents the fully accurate reconstruction of the closed-eye face by our method. Previous works focused on representing the overall features of the facial structure, without paying enough attention to the eye area where some obvious dynamic changes occurred. As shown in Table. \ref{table3} and the cumulative error plot in Fig.\ref{comparison}, our method provides an absolute advantage in accurately reconstructing the 3D face compared with other approaches 3DMM-CNN \cite{3dmm_cnn}, UMDFA \cite{umdfa} and PRNet \cite{prnet}. 

\section{Conclusion}
We introduce a novel ELAM in the context of weakly supervised learning, specifically targeting closed-eye faces. The inclusion of the module significantly enhances the accuracy of reproducing closed-eye states. Additionally, the LDL successfully captures complex details in non-static facial regions, thereby improving the overall realism of the 3D models. Experiments conducted on different datasets further confirms the superiority of our proposed method compared to others. 





 


\vfill\pagebreak


\end{document}